\documentclass[a4,a4wide]{article}
\usepackage{aaai}
\usepackage{times}
\usepackage{helvet}
\usepackage{courier}
\usepackage{graphicx}
\begin{document}
\nocopyright
\title{Nonmonotonic Reasoning  as a Temporal Activity}
\author{Daniel G. Schwartz\\
Department of Computer Science\\
Florida State University\\
Tallahassee, FL 32303\\
}
\maketitle
\begin{abstract}
\begin{quote}
A {\it dynamic reasoning system} (DRS) is an adaptation of a conventional formal logical system that explicitly portrays reasoning as a temporal activity, with each extralogical input to the system and each inference rule application being viewed as occurring at a distinct time step.   Every DRS incorporates some well-defined logic together with a controller that serves to guide the reasoning process in response to user inputs.  Logics are generic, whereas controllers are application-specific.  Every controller does, nonetheless, provide an algorithm for nonmonotonic belief revision.  The general notion of a DRS comprises a framework within which one can formulate the logic and algorithms for a given application and prove that the algorithms are correct, i.e., that they serve to (i) derive all salient information and (ii) preserve the consistency of the belief set.  This paper illustrates the idea with ordinary first-order predicate calculus, suitably modified for the present purpose, and an example.  The example revisits some classic nonmonotonic reasoning puzzles (Opus the Penguin, Nixon Diamond) and shows how these can be resolved in the context of a DRS, using an expanded version of first-order logic that incorporates typed predicate symbols.   All concepts are rigorously defined and effectively computable, thereby providing the foundation for a future software implementation.
\end{quote}
\end{abstract}

\section{1. Introduction}

This paper provide a brief overview of a longer paper that has been accepted for publication, subject to revision, as (Schwartz 2013).  The full text of that paper (64 pages) may be viewed in the arXiv CoRR repository at  http://arxiv.org/abs/1308.5374. 

The notion of a {\it dynamic reasoning system} (DRS) was introduced in (Schwartz 1997) for purposes of formulating reasoning involving a logic of `qualified syllogisms'.  The idea arose in an effort to devise rules for evidence combination.  The logic under study included a multivalent semantics where propositions $P$ were assigned a probabilistic `likelihood value' $l(P)$ in the interval $[0,1]$, so that the likelihood value plays the role of a surrogate truth value.  The situation being modeled is where, based on some evidence, $P$ is assigned a likelihood value $l_1$, and then later, based on other evidence, is assigned a value $l_2$, and it subsequently is desired to combine these values based on some rule into a resulting value $l_3$.  This type of reasoning cannot be represented in a conventional formal logical system with the usual Tarski semantics, since such systems do not allow that a proposition may have more than one truth value; otherwise the semantics would not be mathematically well-defined.  Thus the idea arose to speak more explicitly about different occurrences of the propositions $P$ where the occurrences are separated in time.  In this manner one can construct a well-defined semantics by mapping the different time-stamped occurrences of $P$ to different likelihood/truth values.

In turn, this led to viewing a `derivation path' as it evolves over time as representing the knowledge base, or belief set, of a reasoning agent that is progressively building and modifying its knowledge/beliefs through ongoing interaction with its environment (including inputs from human users or other agents).  It also presented a framework within which one can formulate a Doyle-like procedure for nonmonotonic `reason maintenance' (Doyle 1979; Smith and Kelleher 1988).  Briefly, if the knowledge base harbors inconsistencies due to contradictory inputs from the environment, then in time a contradiction may appear in the reasoning path (knowledge base, belief set), triggering a back-tracking procedure aimed at uncovering the `culprit' propositions that gave rise to the contradiction and disabling (disbelieving) one or more of them so as to remove the inconsistency.  Accordingly the overall reasoning process may be characterized as being `nonmonotonic'. 

Reasoning is nonmonotonic when the discovery and introduction of new information causes one to retract previously held assumptions or conclusions.  This is to be contrasted with classical formal logical systems, which are monotonic in that the introduction of new information (nonlogical axioms) always increases the collection of conclusions (theorems).  (Schwartz 1997) contains an extensive bibliography and survey of the works related to nonmonotonic reasoning as of 1997.  In particular, this includes a discussion of (i) the classic paper by McCarthy and Hayes (McCarthy and Hayes 1969) defining the `frame problem' and describing the `situation calculus', (ii) Doyle's `truth maintenance system' (Doyle1979) and  subsequent `reason maintenance system' (Smith and Kelleher 1988), (iii) McCarthy's `circumscription' (McCarthy 1980), (iv) Reiter's `default logic' (Reiter 1980), and (v) McDermott and Doyle's `nonmonotonic logic' (McDermott and Doyle 1980).  With regard to temporal aspects, there also are discussed works by Shoham and Perlis.  (Shoham 1986; 1988) explores the idea of making time an explicit feature of the logical formalism for reasoning `about' change, and (Shoham 1993) describes a vision of `agent-oriented programming' that is along the same lines of the present DRS, portraying reasoning itself as a temporal activity.  In (Elgot-Drapkin 1988; Elgot-Drapkin et al. 1987; 1991; Elgot-Drapkin and Perlis 1990; Miller 1993; Perlis et al. 1991) Perlis and his students introduce and study the notion of `step logic', which represents reasoning as `situated' in time, and in this respect also has elements in common with the notion of a DRS.  Additionally mentioned but not elaborated upon in (Schwartz 1997) is the so-called AGM framework (Alchour\'on et al. 1985; Gardenfors 1988; 1992), named after its originators.  Nonmonotonic reasoning and belief revision are related in that the former may be viewed as a variety of the latter.

These cited works are nowadays regarded as the classic approaches to nonmonotonic reasoning and belief revision.  Since 1997 the AGM approach has risen in prominence, due in large part to the publication (Hansson 1999), which builds upon and substantially advances the AGM framework.  AGM defines a belief set as a collection of propositions that is closed with respect to the classical consequence operator, and operations of `contraction', `expansion' and `revision' are defined on belief sets.   (Hansson 1999) made the important observation that a belief set can conveniently be represented as the consequential closure of a finite `belief base', and these same AGM operations can be defined in terms of operations performed on belief bases.  Since that publication, AGM has enjoyed a steadily growing population of adherents.  A recent publication (Ferm\'e and Hansson 2011) overviews the first 25 years of research in this area.

The DRS framework has elements in common with AGM, but also differs in several respects.  Most importantly, the present focus is on the creation of computational algorithms that are sufficiently articulated that they can effectively be implemented in software and thereby lead to concrete applications.  This element is still lacking in AGM, despite Hansson's contribution regarding finite belief bases.  The AGM operations continue to be given only as set-theoretic abstractions and have not yet been translated into computable algorithms.

Another research thread that has risen to prominence is the logic-programming approach to nonmonotonic reasoning known as Answer Set Programming (or Answer Set Prolog, aka AnsProlog).  A major work is the treatise (Baral 2003), and a more recent treatment is (Gelfond and Kahl 2014).   This line of research develops an effective approach to nonmonotonic reasoning via an adaptation of the well-known Prolog programming language.  As such, this may be characterized as a `declarative' formulation of nonmonotoniticy, whereas the DRS approach is `procedural'.  The extent to which the two systems address the same problems has yet to be explored. 

A way in which the present approach varies from the original AGM approach, but happens to agree with the views expressed by (Hansson 1999, cf. pp. 15-16), is that it dispenses with two of the original `rationality postulates', namely, the requirements that the underlying belief set be at all times (i) consistent, and (ii) closed with respect to logical entailment.  The latter is sometimes called the `omniscience' postulate, inasmuch as the modeled agent is thus characterized as knowing all possible logical consequences of its beliefs. 

These postulates are intuitively appealing, but they have the drawback that they lead to infinitary systems and thus cannot be directly implemented on a finite computer.  To wit, the logical consequences of even a fairly simple set of beliefs will be infinite in number.  Dropping these postulates does have anthropomorphic rationale, however, since humans themselves cannot be omniscient in the sense described, and, because of this, often harbor inconsistent beliefs without being aware of this.  Thus it is not unreasonable that our agent-oriented reasoning models should have these same characteristics.  Similar remarks may be found in the cited pages of (Hansson 1999). 

Other ways in which the present work differs from the AGM approach may be noted.  First, what is here taken as a `belief set' is neither a belief set in the sense of AGM and Hansson nor a Hansson-style belief base.  Rather it consists of the set of statements that have been input by an external agent as of some time $t$, together with the consequences of those statements that have been derived in accordance with the algorithms provided in a given `controller'.  Second, by labeling the statements with the time step when they are entered into the belief set (either by an external agent or derived by means of an inference rule), one can use the labels as a basis for defining the associated algorithms. Third, whereas G\"ardenfors,  Hansson, and virtually all others that have worked with the AGM framework, have confined their language to be only propositional, the present work takes the next step to full first-order predicate logic.  This is significant inasmuch as the consistency of a finite set of propositions with respect to the classical consequence operation can be determined by truth-table methods, whereas the consistency of a finite set of statements in first-order predicate logic is undecidable (the famous result due to G\"odel).  For this reason the present work develops a well-defined semantics for the chosen logic and establishes a soundness theorem, which in turn can be used to establish consistency.  Last, the present use of a controller is itself new, and leads to a new efficacy for applications.   

The notion of a controller was not present in the previous work (Schwartz 1997).  Its introduction here thus fills an important gap in that treatment.  The original conception of a DRS provided a framework for modeling the reasoning processes of an artificial agent to the extent that those processes follow a well-defined logic, but it offered no mechanism for deciding what inference rules to apply at any given time.  What was missing was a means to provide the agent with a sense of purpose, i.e., mechanisms for pursuing goals. This deficiency is remedied in the present treatment.  The controller responds to inputs from the agent's environment, expressed as propositions in the agent's language.  Inputs are classified as being of various `types', and, depending on the input type, a reasoning algorithm is applied.  Some of these algorithms may cause new propositions to be entered into the belief set, which in turn may invoke other algorithms.  These algorithms thus embody the agent's purpose and are domain-specific, tailored to a particular application.  But in general their role is to ensure that (i) all salient propositions are derived and entered into to the belief set, and (ii) the belief set remains consistent.  The latter is achieved by invoking a Doyle-like reason maintenance algorithm whenever a contradiction, i.e., a proposition of the form $P\land\lnot P$, is entered into the belief set. 

This recent work accordingly represents a rethinking, refinement, and extension of the earlier work, aimed at (1) providing mathematical clarity to some relevant concepts that previously were not explicitly defined, (ii) introducing the notion of a controller and spelling out its properties, and (iii) illustrating these ideas with a small collection of example applications.  As such the work lays the groundwork for a software implementation of the DRS framework, this being a domain-independent software framework into which can be plugged domain-specific modules as required for any given application.  Note that the mathematical work delineated in (Schwartz 2013) is a necessary prerequisite for the software implementation inasmuch as this provides the formal basis for an unambiguous set of requirements specifications.  While the present work employs classical first-order predicate calculus, the DRS framework can accommodate any logic for which there exists a well-defined syntax and semantics.

The following Section 2 provides a fully detailed definition of the notion of a DRS.  Section 3 briefly describes the version of first-order predicate logic introduced for the present purpose and mentions a few items needed for the ensuing discussion. Section 4 illustrates the core ideas in an application to multiple-inheritance systems, showing a new approach to resolving two classic puzzles of nonmontonic reasoning, namely Opus the Penguin and Nixon Diamond.  

\section {2. Dynamic Reasoning Systems}

A {\it dynamic reasoning system} (DRS) comprises a model of an artificial agent's reasoning processes to the extent that those processes adhere to the principles of some well-defined logic.  Formally it is comprised of a `path logic', which provides all the elements necessary for reasoning, and a `controller', which guides the reasoning process. 

\begin{figure}[h]
\centerline{\includegraphics[height=1.5in]{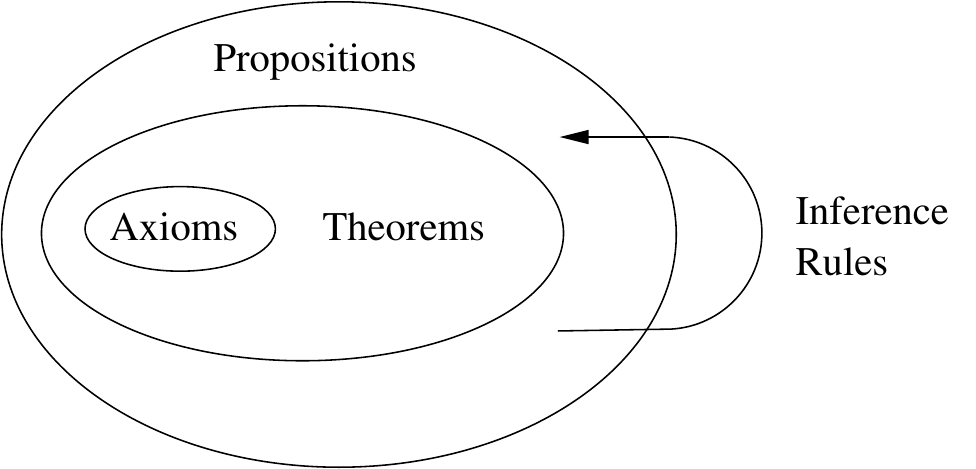}}
\medskip
\caption{Classical formal logical system.}
\end{figure}

\begin{figure}[h]
\centerline{\includegraphics[height=2.0in]{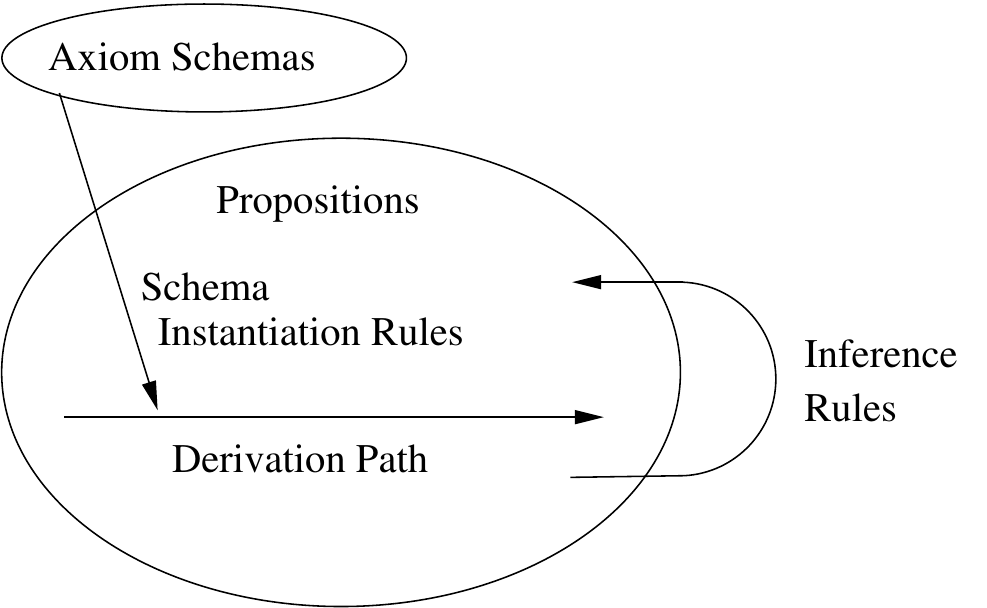}}
\medskip
\caption{Dynamic reasoning system.}
\end{figure}  

For contrast, and by way of introductory overview, the basic structure of a classical formal logical system is portrayed in Figure 1 and that of a DRS in Figure 2.  A classical system is defined by providing a language consisting of a set of propositions, selecting certain propositions to serve as axioms, and specifying a set of inference rules saying how, from certain premises one can derive certain conclusions.  The theorems then amount to all the propositions that can be derived from the axioms by means of the rules.  Such systems are monotonic in that adding new axioms always serves to increase the set of theorems.  Axioms are of two kinds: logical and extralogical (or `proper', or `nonlogical').  The logical axioms together with the inference rules comprise the `logic'. The extralogical axioms comprise information about the application domain.  A DRS begins similarly with specifying a language consisting of a set of propositions.  But here the `logic' is given in terms of a set of axioms schemas, some inference rules as above, and some rules for instantiating the schemas. The indicated derivation path serves as the belief set.  Logical axioms may be entered into the derivation path by applying instantiation rules.  Extralogical axioms are entered from an external source (human user, another agent, a mechanical sensor, etc.).  Thus the derivation path evolves over time, with propositions being entered into the path either as extralogical axioms or derived by means of inference rules in accordance with the algorithms provided in the controller. Whenever a new proposition is entered into the path it is marked as `believed'.  In the event that a contradiction arises in the derivation path, a nonmonotonic belief revision process is invoked which leads to certain previously believed propositions becoming disbelieved, thereby removing the contradiction.  A brief overview of these two components of a DRS is given in Sections 2.1 and 2.2.

\subsection {2.1. Path Logic}

A {\it path logic} consists of a language, axiom schemas, inference rules, and a derivation path, as follows.

{\bf Language}: Here denoted $\cal L$, this consists of all {\it expressions} (or {\it formulas}) that can be generated from a given set $\sigma$ of {\it symbols} in accordance with a collection of production rules (or an inductive definition, or some similar manner of definition).  As symbols typically are of different types (e.g., individual variables, constants, predicate symbols, etc.) it is assumed that there is an unlimited supply (uncountably many if necessary) of each type.  Moreover, as is customary, some symbols will be {\it logical symbols} (e.g., logical connectives, quantifiers, and individual variables), and some will be {\it extralogical symbols} (e.g., individual constants and predicate symbols).  It is assumed that $\cal L$ contains at least the logical connectives for expressing {\it negation} and {\it conjunction}, herein denoted $\lnot$ and $\land$, or a means for defining these connectives in terms of the given connectives.  For example, in the following we take $\lnot$ and $\to$ as given and use the standard definition of $\land$ in terms of these. 

{\bf Axiom Schemas}:  Expressed in some meta notation, these describe the expressions of $\cal L$ that are to serve as {\it logical axioms}.

{\bf Inference Rules}:  These must include one or more rules that enable instantiation of the axiom schemas.  All other inference rules will be of the usual kind, i.e., stating that, from expressions having certain forms (premise expressions), one may infer an expression of some other form (a conclusion expression).  Of the latter, two kinds are allowed: {\it logical rules}, which are considered to be part of the underlying logic, and {\it extralogical rules}, which are associated with the intended application.  Note that logical axioms are expressions that are derived by applying the axiom schema instantiation rules.  Inference rules may be viewed formally as mappings from $\cal L$ into itself.

The rule set may include derived rules that simplify deductions by encapsulating frequently used argument patterns. Rules derived using only logical axioms and logical rules will also be {\it logical rules}, and derived rules whose derivations employ extralogical rules will be additional {\it extralogical rules}. 

{\bf Derivation Paths}: These consist of a sequences of pairs $(L_0,B_0),(L_1,B_1),\ldots$, where $L_t$ is the sublanguage of $\cal L$ that is in use at time $t$, and $B_t$ is the {\it belief set} in effect at time $t$.  Such a sequence is {\it generated} as follows.  Since languages are determined by the symbols they employ, it is useful to speak more directly in terms of the set $\sigma_t$ comprising the symbols that are in use at time $t$ and then let $L_t$ be the sublanguage of $\cal L$ that is based on the symbols in $\sigma_t$.  With this in mind, let $\sigma_0$ be the logical symbols of $\cal L$, so that $L_0$ is the minimal language employing only logical symbols, and let $B_0=\emptyset$. Then, given $(L_t,B_t)$, the pair $(L_{t+1},B_{t+1})$ is formed in one of the following ways: 

\begin{enumerate}

\item $\sigma_{t+1}=\sigma_t$ (so that $L_{t+1}=L_t$) and $B_{t+1}$ is obtained from $B_t$ by adding an expression that is derived by application of an inference rule that instantiates an axiom schema,

\item $\sigma_{t+1}=\sigma_t$ and $B_{t+1}$ is obtained from $B_t$ by adding an expression that is derived from expressions appearing earlier in the path by application of an inference rule of the kind that infers a conclusion from some premises,

\item $\sigma_{t+1}=\sigma_t$ and an expression employing these symbols is added to $B_t$ to form $B_{t+1}$, 

\item some new extralogical symbols are added to $\sigma_t$ to form $\sigma_{t+1}$, and an expression employing the new symbols is added to $B_t$ to form $B_{t+1}$,

\item $\sigma_{t+1}=\sigma_t$ and $B_{t+1}$ is obtained from $B_t$ by applying a belief revision algorithm as described in the following.

\end{enumerate}

Expressions entered into the belief set in accordance with either (3) or (4) will be {\it extralogical axioms}.   A DRS can generate any number of different derivation paths, depending on the extralogical axioms that are input and the inference rules that are applied.

Whenever an expression is entered into the belief set it is assigned a {\it label} comprised of:

\begin{enumerate}

\item A {\it time stamp}, this being the value of the subscript $t+1$ on the set $B_{t+1}$ formed by entering the expression into the belief set in accordance with any of the above items (1) through (4). The time stamp serves as an {\it index} indicating the expression's position in the belief set.

\item A {\it from-list}, indicating how the expression came to be entered into the belief set.  In case the expression is entered in accordance with the above item (1), i.e., using a schema instantiation rule, this list consists of the name (or other identifier) of the schema and the name (or other identifier) of the inference rule if the system has more than one such rule.  In case the expression is entered in accordance with above item (2), the list consists of the indexes (time stamps) of the premise expressions and the name (or other identifier) of the inference rule. In case the expression is entered in accordance with either of items (3) or (4), i.e., is a extralogical axiom, the list will consist of some code indicating this (e.g., {\it es} standing for `external source') possibly together with some identifier or other information regarding the source. 

\item A {\it to-list}, being a list of indexes of all expressions that have been entered into the belief set as a result of rule applications involving the given expression as a premise.  Thus to-lists may be updated at any future time.

\item A {\it status indicator} having the value {\it bel} or {\it disbel} according as the proposition asserted by the expression currently is believed or disbelieved.  The primary significance of this status is that only expressions that are believed can serve as premises in inference rule applications.  Whenever an expression is first entered into the belief set, it is assigned status {\it bel}.  This value may then be changed during belief revision at a later time.  When an expression's status is changed from {\it bel} to {\it disbel} it is said to have been {\it retracted}.

\item An {\it epistemic entrenchment factor}, this being a numerical value indicating the strength with which the proposition asserted by the expression is held. This terminology is adopted in recognition of the work by G\"ardenfors, who initiated this concept (Gardenfors 1988; 1992), and is used here for essentially the same purpose, namely, to assist when making decisions regarding belief retractions.  Depending on the application, however, this value might alternatively be interpreted as a degree of belief, as a certainty factor, as a degree of importance, or some other type of value to be used for this purpose. Logical axioms always receive the highest possible epistemic entrenchment value, whatever scale or range may be employed.

\item A {\it knowledge category specification}, having one of the values {\it a priori}, {\it a posteriori}, {\it analytic}, and {\it synthetic}.  These terms are employed in recognition of the philosophical tradition initiated by Immanuel Kant (Kant 1935).  Logical axioms are designated as a priori; extralogical axioms are designated as a posteriori; expressions whose derivations employ only logical axioms and logical inference rules are designated as analytic; and expressions whose derivations employ any extralogical axioms or extralogical rules are designated as synthetic.
  
\end{enumerate}

Thus when an expression $P$ is entered into the belief set, it is more exactly entered as an expression-label pair $(P,\lambda)$, where $\lambda$ is the label.  A DRS's language, axiom schemas, and inference rules comprise a {\it logic} in the usual sense. It is required that this logic be {\it consistent}, i.e., for no expression $P$ is it possible to derive both $P$ and $\lnot P$. The belief set may become inconsistent, nonetheless, through the introduction of contradictory extralogical axioms.

In what follows, only expressions representing a posteriori and synthetic knowledge may be retracted; expressions of a priori knowledge are taken as being held unequivocally.  Thus the term `a priori knowledge' is taken as synonymous with `belief held unequivocally', and `a posteriori knowledge' is interpreted as `belief possibly held only tentatively' (some a posteriori beliefs may be held unequivocally).  Accordingly the distinction between knowledge and belief is somewhat blurred, and what is referred to as a `belief set' might alternatively be called a `knowledge base', as is often the practice in AI systems.

\subsection {2.2. Controller}
  
A {\it controller} effectively determines the modeled agent's {\it purpose} or {\it goals} by managing the DRS's interaction with its environment and guiding the reasoning process.  With regard to the latter, the objectives typically include (i) deriving all expressions salient to the given application and entering these into the belief set, and (ii) ensuring that the belief set remains consistent.  To these ends, the business of the controller amounts to performing the following operations.

\begin{enumerate}

\item Receiving input from its environment, e.g., human users, sensors, or other artificial agents, expressing this input as expressions in the given language $\cal L$, and entering these expressions into the belief set in the manner described above (derivation path items (3) and (4)).  During this operation, new symbols are appropriated as needed to express concepts not already represented in the current $L_t$. 

\item Applying inference rules in accordance with some extralogical objective (some plan, purpose, or goal) and entering the derived conclusions into the belief set in the manner described above (derivation path items (1) and (2)). 

\item Performing any actions that may be prescribed as a result of the above reasoning process, e.g., moving a robotic arm, returning a response to a human user, or sending a message to another artificial agent.

\item Whenever necessary, applying a `dialectical belief revision' algorithm for contradiction resolution in the manner described below.

\end{enumerate}

A {\it contradiction} is an expression of the form $P\land\lnot P$.  Sometimes it is convenient to represent the general notion of contradiction by the falsum symbol, $\bot$.  Contradiction resolution is triggered whenever a contradiction or a designated equivalent expression is entered into the belief set.  We may assume that this only occurs as the result of an inference rule application, since it obviously would make no sense to enter a contradiction directly as an extralogical axiom.  The contradiction resolution algorithm entails three steps:

\begin{enumerate}

\item Starting with the from-list in the label on the contradictory expression, backtrack through the belief set following from-lists until one identifies all extralogical axioms that were involved in the contradiction's derivation.  Note that such extralogical axioms must exist, since, by the consistency of the logic, the contradiction cannot constitute analytical knowledge, and hence must be synthetic. 

\item Change the belief status of one or more of these extralogical axioms, as many as necessary to invalidate the derivation of the given contradiction.  The decision as to which axioms to retract may be dictated, or at least guided by, the epistemic entrenchment values.  In effect, those expressions with the lower values would be preferred for retraction.  In some systems, this retraction process may be automated, and in others it may be human assisted.

\item Forward chain through to-lists starting with the extralogical axiom(s) just retracted, and retract all expressions whose derivations were dependent on those axioms.  These retracted expressions should include the contradiction that triggered this round of belief revision (otherwise the correct extralogical axioms were not retracted).

\end{enumerate}

This belief revision algorithm is reminiscent of G. W. F. Hegel's `dialectic', described as a process of `negation of the negation' (Hegel 1931).  In that treatment, the latter (first occurring) negation is a perceived internal conflict (here a contradiction), and the former (second occurring) one is an act of transcendence aimed at resolving the conflict (here removing the contradiction).  In recognition of Hegel, the belief revision/retraction process formalized in the above algorithm will be called {\it Dialectical Belief Revision}.

\section{3. First-Order Logic}

The paper (Schwartz 2013) defines a notion of first-order {\it theory} suitable for use in a DRS, provides this with a well-defined semantics (a notion of {\it model\/}), and establishes a Soundness Theorem: a theory is consistent if it has a model.  The notions of theory and semantics are designed to accommodate the notion of a belief set evolving over time, as well as inference rules that act by instantiating axiom schemas.  A first-order language $\cal L$ is defined following the notations of (Hamilton 1988).  This includes notations ${\bf A}_n^m$ as predicate symbols (here the $n$-th $m$-ary predicate symbol) and ${\bf a}_n$ for individual variables. Then, in the path logic, the languages at each successive time step are sublanguages of $\cal L$.  The semantics follows the style of (Shoenfield 1967).  The axiom schemas of (Hamilton 1988) are adopted.  The inference rules are those of (Hamilton 1988) together with some rules for axiom schema instantiation.  The formalism is sufficiently different from the classical version that new proofs of all relevant propositions must be restated in this context and proven correct.  The treatment also establishes the validity of several derived inference rules that become useful in later examples, including: 

\begin{description}

\item[\kern 2em{\bf Hypothetical Syllogism}] From $P\to Q$ and $Q\to R$ infer $P\to R$, where $P,Q,R$ are any formulas.

\item[\kern 2em {\bf Aristotelian Syllogism}] From $(\forall x)(P\to Q)$ and $P(a/x)$, infer $Q(a/x)$, where $P,Q$ are any formulas, $x$ is any individual variable, and $a$ is any individual constant.

\item[\kern 2em{\bf Subsumption}] From $(\forall x)(\alpha(x)\to\beta(x))$ and $(\forall x)(\beta(x)\to\gamma(x))$, infer $(\forall x)(\alpha(x)\to\gamma(x))$, where $\alpha,\beta,\gamma$ are any unary predicate symbols, and $x$ is any individual variable.

\item[\kern 2em{\bf Contradiction Detection}] From $P$ and $\lnot P$ infer $\bot$, where $P$ is any formula.

\item[\kern 2em{\bf Conflict Detection}] From $(\forall x)\lnot(P\land Q)$, $P(a/x)$, and $Q(a/x)$ infer $\bot$, where $P,Q$ are any formulas, $x$ is any individual variable, and $a$ is any individual constant.

\end{description}  

\section {4. Example: Multiple Inheritance with Exceptions}

The main objective of (Schwartz 1997) was to show how a DRS framework could be used to formulate reasoning about property inheritance with exceptions, where the underlying logic was a probabilistic `logic of qualified syllogisms'.  This work was inspired in part by the frame-based systems due to (Minsky 1975) and constitutes an alternative formulation of the underlying logic (e.g., as discussed by (Hayes 1980)).
   
What was missing in (Schwartz 1997) was the notion of a controller.  There a reasoning system was presented and shown to provide intuitively plausible solutions to numerous `puzzles' that had previously appeared in the literature on nonmonotonic reasoning, e.g., Opus the Penguin (Touretsky 1984), Nixon Diamond (Touretsky et al. 1987), and Clyde the Elephant (Touretsky et al. 1987).  But there was nothing to guide the reasoning processes---no means for providing a sense of purpose for the reasoning agent.  The present work fills this gap by adding a controller. Moreover, it deals with a simpler system based on first-order logic and remands further exploitation of the logic of qualified syllogisms to a later work.  The kind of DRS developed in this section will be termed a {\it multiple inheritance system} (MIS).

For this application the language $\cal L$  discussed in Section 3 is expanded by including some {\it typed predicate symbols}, namely, some unary predicate symbols ${\bf A}^{(k)}_1,{\bf A}^{(k)}_2,\ldots$  representing {\it kinds} of things (any objects), and some unary predicate symbols ${\bf A}^{(p)}_1,{\bf A}^{(p)}_2,\ldots$ representing {\it properties} of things.  The superscripts $k$ and $p$ are applied also to generic denotations.  Thus an expression of the form $(\forall x)(\alpha^{(k)}(x)\to\beta^{(p)}(x))$ represents the proposition that all $\alpha$s have property $\beta$.  These new predicate symbols are used here purely as syntactical items for purposes of defining an extralogical `specificity principle' and some associated extralogical graphical structures and algorithms.  Semantically they are treated exactly the same as other predicate symbols.
  
A {\it multiple-inheritance hierarchy} $H$ will be a directed graph consisting of a set of {\it nodes} together with a set of {\it links} represented as ordered pairs of nodes.  Nodes may be either {\it object} nodes, {\it kind} nodes, or {\it property} nodes.  A link of the form (object node, kind node) will be an {\it object-kind\/} link, one of the form (kind node, kind node) will be a {\it subkind-kind\/} link, and one of the form (kind node, property node) will be a {\it has-property\/} link. There will be no other types of links.  Object nodes will be labeled with (represent) individual constant symbols, kind nodes will be labeled with (represent) kind-type unary predicate symbols, and property nodes will be labeled with (represent) property-type unary predicate symbols or negations of such symbols.  In addition, each property type predicate symbol with bear a numerical subscript, called an {\it occurrence index}, indicating an occurrence of that symbol in a given hierarchy $H$.  These indexes are used to distinguish different occurrences of the same property-type symbol in $H$.   An object-kind link between an individual constant symbol $a$ and a  predicate symbol $\alpha^{(k)}$ will represent the formula $\alpha^{(k)}(a)$, a subkind-kind link between  a predicate symbol $\alpha^{(k)}$ and a predicate symbol $\beta^{(k)}$ will represent the formula $(\forall x)(\alpha^{(k)}(x)\to\beta^{(k)}(x))$, and a has-property link between a predicate symbol  $\alpha^{(k)}$ and a predicate symbol $\beta^{(p)}_1$ will represent the formula $(\forall x)(\alpha^{(k)}(x)\to\beta^{(p)}_1(x))$.  

Given such an $H$, there is defined on the object nodes and the  kind nodes a {\it specificity relation} $>_s$ (read `more specific than') according to:  (i) if $({\rm node}_1,{\rm node}_2)$ is either an object-kind link or a kind-kind link, then ${\rm node}_1>_s{\rm node}_2$, and (ii) if ${\rm node}_1>_s{\rm node}_2$ and ${\rm node}_2>_s{\rm node}_3$, then ${\rm node}_1>_s{\rm node}_3$.  We shall also have a dual {\it generality relation} $>_g$ (read `more general than') defined by ${\rm node}_1>_g{\rm node}_2$ iff ${\rm node}_2>_s{\rm node}_1$.  It follows that object nodes are maximally specific and minimally general.  It also follows that $H$ may have any number of maximally general nodes, and in fact that it need not be connected.  A maximally general node is a {\it root} node.  A {\it path} in a hierarchy $H$ (not to be confused with the path in a path logic) will be a sequence ${\rm node}_1,\ldots,{\rm node}_n$ wherein, ${\rm node}_1$ is a root node and, for each $i=1,\ldots,n-2$, the pair $({\rm node}_{i+1},{\rm node}_i)$ is a subkind-kind link, and, the pair $({\rm node}_n,{\rm node}_{n-1})$ is either a subkind-kind link or an object-kind link. Note that property nodes do not participate in paths as here defined.  

It is desired to organize a multiple inheritance hierarchy as a directed acyclic graph (DAG) without redundant links with respect to the object-kind and subkind-kind links (i.e., here ignoring has-property links), where, as before, by a redundant link is meant a direct link from some node to an ancestor of that node other than the node's immediate ancestors (i.e., other than its parents).  More exactly, two distinct paths will form a {\it redundant pair} if they have some node in common beyond the first place where they differ.  This means that they comprise two distinct paths to the common node(s).  A path will be simply {\it redundant} (or {\it redundant in} $H$) if it is a member of a redundant pair.  A path contains a {\it loop} if it has more than one occurrence of the same node.  Provisions are made in the following algorithms to ensure that hierarchies with loops or redundant paths are not allowed.  As is customary, the hierarchies will be drawn with the upward direction being from more specific to less (less general to more), so that roots appear at the top and objects appear at the bottom.  Kind-property links will extend horizontally from their associated kind nodes.

In terms of the above specificity relation on $H$, we can assign an {\it address} to each object and kind node in the following manner.  Let the addresses of the root nodes, in any order, be $(1),(2),(3),\ldots$.  Then for the node with address (1), say, let the next most specific nodes in any order have the addresses $(1,1),(1,2),(1,3),\ldots$; let the nodes next most specific to the one with address $(1,1)$ have addresses $(1,1,1),(1,1,2),(1,1,3),\ldots$; and so on.  Thus an address indicates the node's position in the hierarchy relative to some root node.  Inasmuch as an object or kind node may be more specific than several different root nodes, the same node may have more than one such address.  Note that the successive initial segments of an address are the addresses of the nodes appearing in the path from the related root node to the node having that initial segment as its address.  Let $>$ denote the usual lexicographic order on addresses.  We shall apply $>$ also to the nodes having those addresses.  It is easily verified that, if ${\rm node}_1>{\rm node}_2$  and the ${\rm node}_2$ address is an initial segment of the ${\rm node}_1$ address, then ${\rm node}_1>_s{\rm node}_2$, and conversely.  For object and kind nodes, we shall use the term {\it specificity rank} (or just {\it rank}) synonymously with `address'. 

Since, as mentioned, it is possible for any given object or kind node to have more than one address, it thus can have more than one rank.  Two nodes are comparable with respect to the specificity relation $>_s$, however, only if they appear on the same path, i.e., only if one node is an ancestor of the other, in which case only the rank each has acquired due to its being on that path will apply.  Thus, if two nodes are comparable with respect to their ranks by the relation $>_s$, there is no ambiguity regarding the ranks being compared.

Having thus defined specificity ranks for object and kind nodes, let us agree that each property node inherits the rank of the kind node to which it is linked.  Thus for property nodes the rank is not an address.

\begin{figure}[htp]
\centerline{\includegraphics[height=1.55in]{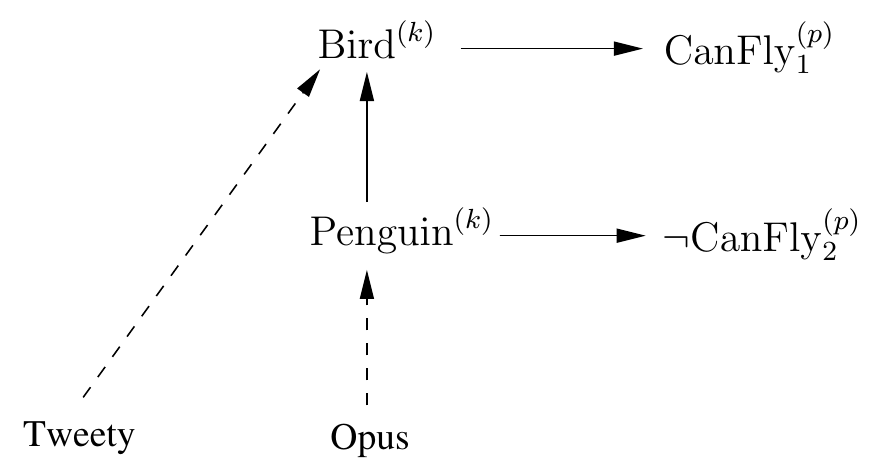}}
\medskip
\caption{Tweety the Bird and Opus the Penguin as an MIS.}
\end{figure}  

An example of such a hierarchy is shown in Figure 3.  Here `Tweety' and `Opus' may be taken as names for the individual constants ${\bf a}_1$ and ${\bf a}_2$, and `${\rm Bird}^{(k)}$', `${\rm Penguin}^{(k)}$', and `${\rm CanFly}^{(p)}$' can be taken as names, respectively, for the unary predicate symbols ${\bf A}^{(k)}_1$,  ${\bf A}^{(k)}_2$, and ${\bf A}^{(p)}_1$.  [Note: The superscripts are retained on the names only to visually identify the types of the predicate symbols, and could be dropped without altering the meanings.] The links represent the formulas \medskip

\indent\indent $(\forall x)({\rm Penguin}^{(k)}(x)\to{\rm Bird}^{(k)}(x))$

\indent\indent $(\forall x)({\rm Bird}^{(k)}(x)\to{\rm CanFly}^{(p)}_1(x))$

\indent\indent $(\forall x)({\rm Penguin}^{(k)}(x)\to\lnot{\rm CanFly}^{(p)}_2(x))$

\indent\indent ${\rm Bird}^{(k)}({\rm Tweety})$

\indent\indent ${\rm Penquin}^{(k)}({\rm Opus})$  \medskip

\noindent The subscripts 1 and 2 on the predicate symbol ${\rm CanFly}^{(p)}$ in the graph distinguish the different occurrences of this symbol in the graph, and the same subcripts on the symbol occurrences in the formulas serve to correlate these with their occurrences in the graph.  Note that these are just separate occurrences of the same symbol, however, and therefore have identical semantic interpretations.  Formally,  ${\rm CanFly}^{(p)}_1$ and ${\rm CanFly}^{(p)}_2$ can be taken as standing for ${\bf A}^{(p)}_{1_1}$ and ${\bf A}^{(p)}_{1_2}$ with the lower subscripts being regarded as extralogical notations indicating different occurrences of ${\bf A}^{(p)}_1$.

This figure reveals the rationale for the present notion of multiple-inheritance hierarchy.  The intended interpretation of the graph is that element nodes and kind nodes inherit the properties of their parents, with the exception that more specific property nodes take priority and block inheritances from those that are less specific.  Let us refer to this as the {\it specificity principle}. In accordance with this principle, in Figure 3 Tweety inherits the property CanFly from Bird, but Opus does not inherit this property because the inheritance is blocked by the more specific information that Opus is a Penguin and Penguins cannot fly.  

\begin{figure}[htp]
\centerline{\includegraphics[height=1.75in]{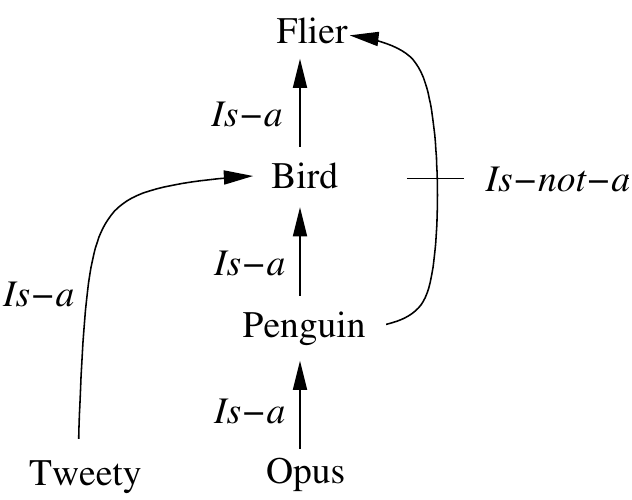}}
\medskip
\caption{Tweety the Bird and Opus the Penguin, original version.}
\end{figure}  

Figure 3 constitutes a rethinking of the well-known example of Opus the penguin depicted in Figure 4 (adapted from (Touretsky1984)).  The latter is problematic in that, by one reasoning path one can conclude that Opus is a flier, and by another reasoning path that he is not. This same contradiction is implicit in the formulas introduced above, since if one were to apply the axioms and rules of first-order logic discussed in Section 3, one could derive both ${\rm CanFly}^{(p)}({\rm Opus})$ and $\lnot{\rm CanFly}^{(p)}({\rm Opus})$, in which case the system would be inconsistent.  

\subsection{Formal Specification of an Arbitrary MIS}

We are now in a position to define the desired kind of DRS.  For the path logic, let the language be the one described above, obtained from the $\cal L$ of Section 3 by adjoining the additional unary kind-type and property-type predicate symbols, let the axiom schemas and inference rules be those discussed in Section 3 together with Aristotelian Syllogism and Contradiction Detection.  In this case, derivation paths will consist of triples $(L_t,B_t,H_t)$, where these components respectively are the (sub)language (of $\cal L$), belief set, and multiple inheritance hierarchy at time $t$.  In accordance with Section 2, let $L_0$ be the minimal sublanguage of $\cal L$ consisting of all formulas that can be built up from the atomic formula $\bot$, and let $B_0=\emptyset$.  In addition, let $H_0=\emptyset$.

The MIS controller is designed to enforce the above specificity principle.  Contradictions can arise in an MIS that has inherently contradictory root nodes in its multiple inheritance hierarchy.  An example of this, the famous Nixon Diamond (Touretsky 1987), will be discussed. The purpose of the MIS controller will be (i) to derive and enter into the belief set all object classifications implicit in the multiple inheritance hierarchy, i.e., all formulas of the form $\alpha^{(k)}(a)$ that can be derived from formulas describing the hierarchy (while observing the specificity principle), and (ii) to ensure that the belief set remains consistent. Item (i) thus defines what will be considered the {\it salient information} for an MIS.  Also, the MIS controller is intended to maintain the multiple inheritance hierarchy as a DAG without redundant paths with respect to just the object and kind nodes.  Formulas that can be input by the users may have one of the forms (i) $\alpha^{(k)}(a)$, (ii) $(\forall x)(\alpha^{(k)}(x)\to\beta^{(k)}(x))$, (iii)  $(\forall x)(\alpha^{(k)}(x)\to\beta^{(p)}(x))$, and (iv)  $(\forall x)(\alpha^{(k)}(x)\to\lnot\beta^{(p)}(x))$.   It will be agreed that the epistemic entrenchment value for all input formulas is $0.5$.

We may now define some algorithms that are to be executed in response to each type of user input.  There will be eight types of events.  Event Types 1, 6, 7 and 8 correspond to user inputs, and the others occur as the result of rule applications.  In all such events it is assumed that, if the formula provided to the controller already exists and is active in the current belief set, its input is immediately rejected.  In each event, assume that the most recent entry into the derivation path is $(L_t,B_t,H_t)$. For the details of the algorithms, please see (Schwartz 2013). \medskip

{\bf Event Type 1:}  A formula of the form $\alpha^{(k)}(a)$ is provided to the controller by a human user.  \medskip

{\bf Event Type 2:} A formula of the form $\alpha^{(k)}(a)$ is provided to the controller as a result of an inference rule application (Aristotelian Syllogism). \medskip

{\bf Event Type 3:} A formula of the form $\alpha^{(p)}(a)$ is provided to the controller as a result of an inference rule application (Aristotelian Syllogism).  \medskip

{\bf Event Type 4:} A formula of the form $\lnot\alpha^{(p)}(a)$ is provided to the controller as a result of an inference rule application (Aristotelian Syllogism).   \medskip

{\bf Event Type 5:} The formula $\bot$ is provided to the controller as the result of an application of Contradiction Detection.  \medskip

{\bf Event Type 6:}  A formula of the form $(\forall x)(\alpha^{(k)}(x)\to\beta^{(k)}(x))$ is provided to the controller by a human user.  \medskip

{\bf Event Type 7:}   A formula of the form $(\forall x)(\alpha^{(k)}(x)\to\beta^{(p)}(x))$ is provided to the controller by a human user.  \medskip

{\bf Event Type 8:}   A formula of the form $(\forall x)(\alpha^{(k)}(x)\to\lnot\beta^{(p)}(x))$ is provided to the controller by a human user.   \medskip

\section{Main Results}

That an MIS controller produces all relevant salient information as prescribed above can be summarized as a pair of theorems.  \medskip

{\bf Theorem 5.1.} The foregoing algorithms serve to maintain the hierarchy with respect to the object and kind nodes as a directed acyclic graph without redundant links. \medskip

{\bf Theorem 5.2.} After any process initiated by a user input terminates, the resulting belief set will contain a formula of the form $\alpha^{(k)}(a)$ or $\alpha^{(p)}(a)$ or $\lnot\alpha^{(p)}(a)$  iff the formula is derivable from the formulas corresponding to links in the inheritance hierarchy, observing the specificity principle.  \medskip 

That the algorithms serve to preserve the consistency of the belief set is established as: \medskip

{\bf Theorem 5.3.} For any derivation path in an MIS, the belief set that results at the conclusion of a process initiated by a user input will be consistent with respect to the formulas of the forms $\alpha^{(k)}(a)$, $(\forall x)(\alpha^{(k)}(x)\to\beta^{(p)}(x))$, and $\alpha^{(p)}(a)$.  \medskip

\section {Illustration 1}

Some of the algorithms associated with the foregoing events can be illustrated by considering the inputs needed to create the inheritance hierarchy shown in Figure 3.  This focuses on the process of property inheritance with exceptions.  Let us abbreviate `Bird', `Penguin', and `CanFly', respectively, by `B', `P', and `CF'.   In accordance with the definition of derivation path in Section 2.1, the language $L_0$ will consist only of the formula $\bot$, and the belief set $B_0=\emptyset$.  In accordance with the definition of an MIS, $H_0=\emptyset$.  We consider inputs of the afoermentioned formulas, with each input comprising a type of event initiating a particular reasoning algorithm.  These inputs and event types  are: \medskip

$(\forall x)({\rm P}^{(k)}(x)\to{\rm B}^{(k)}(x))$, Type 6 

$(\forall x)({\rm B}^{(k)}(x)\to{\rm CF}^{(p)}_1(x))$, Type 7

$(\forall x)({\rm P}^{(k)}(x)\to\lnot{\rm CF}^{(p)}_2(x))$, Type 8 

${\rm B}^{(k)}({\rm Tweety})$, Type 1

${\rm P}^{(k)}({\rm Opus})$,  Type 1 \medskip

The specificity principle is invoked during the last event.  This results in the following belief set (omitting formula labels): \medskip

 $(\forall x)({\rm P}^{(k)}(x)\to{\rm B}^{(k)}(x))$

$(\forall x)({\rm B}^{(k)}(x)\to{\rm CF}^{(p)}_1(x))$

$(\forall x)({\rm P}^{(k)}(x)\to\lnot{\rm CF}^{(p)}_2(x))$

${\rm B}^{(k)}({\rm Tweety})$

${\rm CF}^{(p)}_1({\rm Tweety})$

${\rm P}^{(k)}({\rm Opus})$

${\rm B}^{(k)}({\rm Opus})$

$\lnot{\rm CF}^{(p)}_2({\rm Opus})$ \medskip

Thus is is seen that, in this example, the algorithms serve to derive all salient information, i.e., all formulas of the forms $\alpha^{(k)}(a)$, $\alpha^{(p)}(a)$, and $\alpha^{(p)}(a)$ that are implicit in the graph, while at the same time correctly enforcing the specificity principle.  It may also be observed that the belief set is consistent.

\section {Illustration 2}

This considers an application of Contradiction Detection.  The classic Nixon Diamond puzzle (cf. Touretsky et al. 1987) is shown in Figure 5.  Here a contradiction arises because, by the reasoning portrayed on the left side, Nixon is a pacifist, whereas, by the reasoning portrayed on the right, he is not.  The resolution of this puzzle in the context of an MIS can be described in terms of the multiple inheritance hierarchy shown in Figure 6.

\begin{figure}[htp]
\centerline{\includegraphics[height=1.9in]{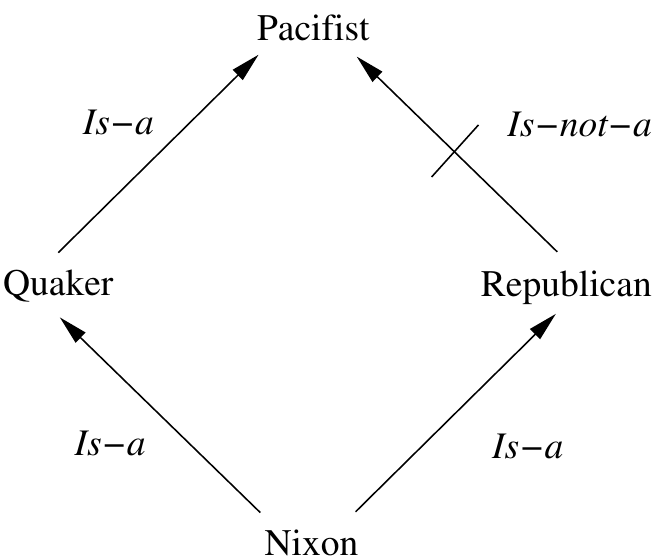}}
\medskip
\caption{Nixon Diamond, original version.}
\end{figure}  

\begin{figure}[htp]
\centerline{\includegraphics[height=1in]{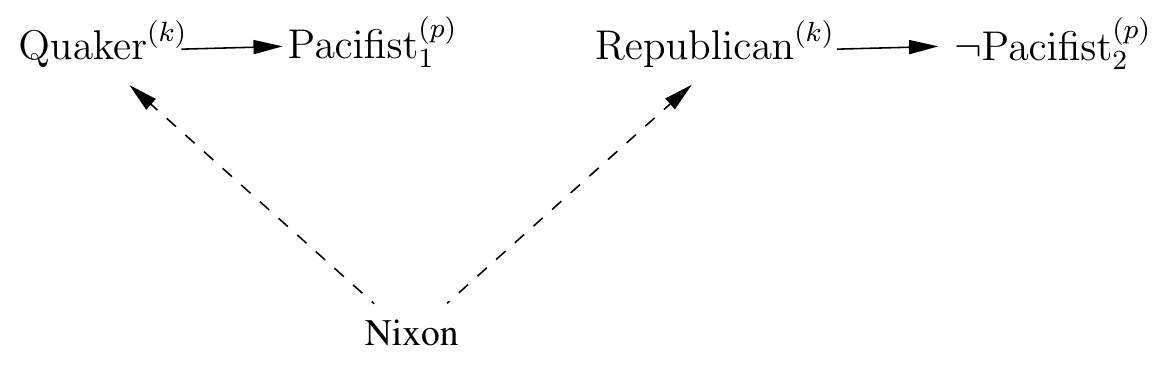}}
\medskip
\caption{Nixon Diamond as an MIS.}
\end{figure}  

The links in Figure 6 represent the formulas \medskip

\indent\indent $(\forall x)({\rm Quaker}^{(k)}(x)\to{\rm Pacifist}^{(p)}_1(x))$

\indent\indent $(\forall x)({\rm Republican}^{(k)}(x)\to\lnot{\rm Pacifist}^{(p)}_2(x))$

\indent\indent ${\rm Quaker}^{(k)}({\rm Nixon})$

\indent\indent ${\rm Republican}^{(k)}({\rm Nixon})$ \medskip

The action of the algorithms may be traced similarly as in Illustration 1.  Let `Quaker',  `Republican' and `Pacifist' denote the predicate symbols ${\bf A}^{(k)}_1$, ${\bf A}^{(k)}_2$ and ${\bf A}^{(p)}_1$, and abbreviate these by `Q', `R' and `P'.  Let `Nixon' denote the individual constant ${\bf a}_1$. $L_0$, $B_0$, and $H_0$ will be as before.  The inputs and their event types are: \medskip

$(\forall x)({\rm Q}^{(k)}(x)\to{\rm P}^{(p)}_1(x))$, Type 7. 

$(\forall x)({\rm R}^{(k)}(x)\to\lnot{\rm P}^{(p)}_1(x))$, Type 8.

${\rm Q}^{(k)}({\rm Nixon})$, Type 1. 

${\rm R}^{(k)}({\rm Nixon})$, Type 1.  \medskip

These lead to the following belief set (again omitting formual labels): \medskip

 $(\forall x)({\rm Q}^{(k)}(x)\to{\rm P}^{(p)}_1(x))$

 $(\forall x)({\rm R}^{(k)}(x)\to\lnot{\rm P}^{(p)}_2(x))$

${\rm Q}^{(k)}({\rm Nixon})$. 

${\rm P}^{(p)}_1({\rm Nixon})$

${\rm R}^{(k)}({\rm Nixon}$

$\lnot{\rm P}^{(p)}_2({\rm Nixon})$

 $\bot$ \medskip

At this point Dialectical Belief Revision is invoked.  All the formulas that were input by the user are candidates for belief change.  Suppose that the formula $(\forall x)({\rm R}^{(k)}(x)\to\lnot{\rm P}^{(p)}_2(x))$, is chosen.  Then the procedure forward chains through to lists, starting with this formula, and changes to {\it disbel} the status first of $\lnot{\rm P}^{(p)}_2({\rm Nixon})$, and then of $\bot$. This results in a belief set with these three formulas removed (disbelieved)  leaving only the left side of the hierarchy in Figure 6.  Thus again all salient information is derived and the resulting belief set is consistent.       

Further well-known puzzles that can be resolved similarly within an MIS are the others discussed in (Schwartz 1997), namely, Bosco the Blue Whale (Stein 1992), Suzie the Platypus (Stein 1992), Clyde the Royal Elephant (Touretsky et al. 1987), and Expanded Nixon Diamond  (Touretsky et al. 1987).

\section{References}

%\begin{thebibliography}{1}

\smallskip \noindent Alchour\'on, C. E.; G\"ardenfors, P.; and Makinson, D. 1985. On the logic of theory change: partial meet contraction and revision
  functions. {\it Journal of Symbolic Logic} 50(2):510--530.

\smallskip \noindent Baral, C. 2003. {\it Knowledge Representation, Reasoning, and Declarative Problem
  Solving}. Cambridge University Press.

\smallskip \noindent Delgrande, J. P., and Farber, W., eds. 2011. {\it Logic Programming and Nonmonotonic Reasoning 11th International
  Conference, LPNMR 2011}.  Lecture notes in Computer Science, Volume 66452011, Springer Verlag.

\smallskip \noindent Doyle, J. 1979. A truth maintenance system. {\it Artificial Intelligence} 12:231--272.

\smallskip \noindent Elgot-Drapkin, J. J. 1988. {\it Step Logic: Reasoning Situated in Time}.  PhD thesis, University of Maryland, College Park. Technical Report CS-TR-2156 and UMIACS-TR-88-94.

\smallskip \noindent Elgot-Drapkin, J. J.; Miller, M.; and Perlis, D.  1987.  Life on a desert island: ongoing work on real-time reasoning. In F.M. Brown, ed., {\it The Frame Problem in Artificial Intelligence: Proceedings of the 1987 Workshop}, pp. 349--357, Los Altos, CA: Morgan Kaufmann.

\smallskip \noindent Elgot-Drapkin, J. J.; Miller, M.; and Perlis, D. 1991. Memory, reason, and time: the step-logic approach. In R.~Cummins and J.~Pollock, eds, {\it Philosophy and AI: Essays at the Interface}, pp. 79--103.  MIT Press.

\smallskip \noindent Elgot-Drapkin, J. J., and Perlis, D. 1990.  Reasoning situated in time I: basic concept. {\it Journal of Experimental and Theoretical Artificial Intelligence} 2(1):75--98.

\smallskip \noindent Gelfond, M. and Kahl, Y., {\it Knowledge Representation, Reasoning, and the Design of  Intelligent Agents: The Anwer Set Programming Approach}, Cambridge University Press, 2014. 

\smallskip \noindent Hayes, P. J. 1980. The logic of frames.  In D.  Metzing, ed., {\it Frame Conceptions and Text Understanding}, Berlin: Walter de Gruyter, pp. 46--61.

\smallskip \noindent Ferm\'e, E., and Hansson, S. O. 2011.  AGM 25 years: twenty-five years of research in belief change. {\it J. Philos Logic}, 40:295--331.

\smallskip \noindent G\"ardenfors, P.  1988.  {\it Knowledge in Flux: Modeling the Dynamics of Epistemic States}. Cambridge, MA: MIT Press/Bradford Books.

\smallskip \noindent G\"ardenfors, P., ed. 1992. {\it Belief Revision}. Cambridge University Press.

\smallskip \noindent  Ginsberg, M. L., ed. 1987. {\it Readings in Nonmonotonic Reasoning}. Los Altos, CA: Morgan Kaufmann.

\smallskip \noindent  Hamilton, A. G. 1988. {\it Logic for Mathematicians, Revised Edition}, Cambridge University Press.

\smallskip \noindent Hansson, S.O. 1999. {\it A Textbook of Belief Dynamics: Theory Change and Database Updating}. Dordercht, Kluwer Academic Publishers.

\smallskip \noindent Hegel, G.W.F. 1931. {\it Phenomenology of Mind}. J.B. Baillie, trans, 2nd edition.  Oxford: Clarendon Press.

\smallskip \noindent Kant, I. 1935 {\it Critique of Pure Reason}. N.K. Smith, trans. London, England: Macmillan.

\smallskip \noindent McCarthy, J. 1980. Circumscription---a form of nonmonotonic reasoning.  {\it Artificial Intelligence}, 13:27--39, 171--172.
Reprinted in (Ginsberg 1987), pp. 145--152.

\smallskip \noindent McCarthy, J., and Hayes, P. 1969. Some philosophical problems from the standpoint of artificial intelligence.  Stanford University. Reprinted in (Ginsberg 1987), pp. 26--45, and in V. Lifschitz, ed., {\it Formalizing Common Sense: Papers by John McCarthy}, Norwood,
  NJ: Ablex, 1990, pp. 21--63.

\smallskip \noindent McDermott, D., and Doyle, J. 1980.  Non-monotonic logic--I. {\it Artificial Intelligence} 13:41--72.  Reprinted in (Ginsberg 1987), pp. 111--126.

\smallskip \noindent Miller, M. J. 1993. {\it A View of One's Past and Other Aspects of Reasoned Change in Belief}.  PhD thesis, University of Maryland, College Park, Department of Computer Science, July. Technical Report CS-TR-3107 and UMIACS-TR-93-66.

\smallskip \noindent Minsky, M. 1975.  A framework for representing knowledge.  In  P. Winston, ed., {\it The Psychology of Computer Vision}, New York: McGraw-Hill, pp. 211--277.  A condensed version has appeared in  D. Metzing, ed., {\it Frame Conceptions and Text Understanding}, Berlin: Walter de
Gruyter, Berlin, 1980, pp. 1--25.

\smallskip \noindent Perlis, D.; Elgot-Drapkin, J. J.; and Miller, M. 1991.  Stop the world---I want to think.  In K.~Ford and F.~Anger, eds., {\it International Journal of Intelligent Systems: Special Issue on Temporal Reasoning, Vol. 6}, pp. 443--456.  Also Technical Report CS-TR-2415 and UMIACS-TR-90-26, Department of Computer Science, University of Maryland, College Park, 1990.

\smallskip \noindent Reiter, R. 1980.  A logic for default reasoning. {\it Artificial Intelligence} 13(1-2):81--132. Reprinted in (Ginsberg 1987), pp. 68--93.

\smallskip \noindent Schwartz, D. G. 1997. Dynamic reasoning with qualified syllogisms. {\it Artificial Intelligenc} 93:103--167.

\smallskip \noindent Schwartz, D .G. 2013. Dynamic reasoning systems. {\it ACM Transactions on Computational Intelligence}, accepted subject to revision February 7, 2014.

\smallskip \noindent Shoenfield, J. R. 1967. {\it Mathematical Logic}, Association for Symbolic Logic.

\smallskip \noindent Shoham, Y. 1986. Chronological ignorance: time, nonmonotonicity, necessity, and causal theories. {\it Proceedings of the American Association for Artificial Intelligence, AAAI'86}, Philadelphia, PA, pp. 389--393.

\smallskip \noindent Shoham, Y. 1988. {\it Reasoning about Change: Time and Causation from the Standpoint of Artificial Intelligence}. Cambridge, MA: MIT Press.

\smallskip \noindent Shoham, Y. 1993. Agent-oriented programming. {\it Artificial Intelligence} 60:51--92.

\smallskip \noindent Smith, B., and Kelleher, G., eds. 1988. {\it Reason Maintenance Systems and Their Applications}.  Chichester, England:Ellis Horwood.

\smallskip \noindent Stein, L. A. 1992. Resolving ambiguity in nonmonotonic inheritance hierarchies. {\it Artificial Intelligence} 55(2-3).

\smallskip \noindent Touretzky, D.  1984. Implicit ordering of defaults in inheritance systems. {\it Proceedings of the Fifth National Conference on Artificial
  Intelligence, AAAI'84}, Austin, TX, Los Altos, CA: Morgan Kaufmann,  pp. 322--325. Reprinted in (Ginsberg 1987), pp. 106--109, and in G. Shafer and J. Pearl, eds., {\it Readings in Uncertain Reasoning}, San Mateo, CA: Morgan Kaufmann, 1990, pp. 668--671.

\smallskip \noindent Touretzky, D. S.; Horty, J .E.; and Thomason, R.H. 1987. A clash of intuitions: the current state of nonmonotonic multiple
  inheritance systems.  {\it Proceedings of the International Joint Conference on Artificial Intelligence, IJCAI'87}, Milan, Italy. pp. 476--482.

\end{document}